\documentclass[letterpaper, 10 pt, conference]{ieeeconf}  % Comment this line out if you need a4paper
\usepackage{xcolor}

\IEEEoverridecommandlockouts                              % This command is only needed if 
\usepackage{siunitx}
\usepackage{graphics} % for pdf, bitmapped graphics files
\usepackage{epsfig} % for postscript graphics files
\usepackage{mathptmx} % assumes new font selection scheme installed
\usepackage{times} % assumes new font selection scheme installed
\usepackage{amsmath} % assumes amsmath package installed
\usepackage{amssymb}  % assumes amsmath package installed
\usepackage{lipsum}

\title{\LARGE \bf
Embodied multi-modal sensing with a soft modular arm powered by physical reservoir computing\\
}

\author{Jun Wang$^{1}$ and Suyi Li$^{1}$% <-this % stops a space
\thanks{*This work was supported by National Science Foundation}% <-this % stops a space
\thanks{$^{1}$Jun Wang and Suyi Li is with the Department of Mechanical Engineering, Virginia Tech, 181 Durham Hall, 1145 Perry Street, Blacksburg, VA 24061, USA
        {\tt\small (email:{junw,suyili}@vt.edu)}}%
}

\begin{document}

\maketitle
\thispagestyle{empty}
\pagestyle{empty}

%%%%%%%%%%%%%%%%%%%%%%%%%%%%%%%%%%%%%%%%%%%%%%%%%%%%%%%%%%%%%%%%%%%%%%%%%%%%%%%%
\begin{abstract}

Soft robots have become increasingly popular for complex manipulation tasks requiring gentle and safe contact. However, their softness makes accurate control challenging, and high-fidelity sensing is a prerequisite to adequate control performance. To this end, many flexible and embedded sensors have been created over the past decade, but they inevitably increase the robot's complexity and stiffness. This study demonstrates a novel approach that uses simple bending strain gauges embedded inside a modular arm to extract complex information regarding its deformation and working conditions. The core idea is based on physical reservoir computing (PRC): A soft body's rich nonlinear dynamic responses, captured by the inter-connected bending sensor network, could be utilized for complex multi-modal sensing with a simple linear regression algorithm. Our results show that the soft modular arm reservoir can accurately predict body posture (bending angle), estimate payload weight, determine payload orientation, and even differentiate two payloads with only minimal difference in weight --- all using minimal digital computing power. 

\end{abstract}

%%%%%%%%%%%%%%%%%%%%%%%%%%%%%%%%%%%%%%%%%%%%%%%%%%%%%%%%%%%%%%%%%%%%%%%%%%%%%%%%

\section{Introduction}

\begin{figure}[hbt]
    \centering
    \includegraphics[width=1.0\linewidth]{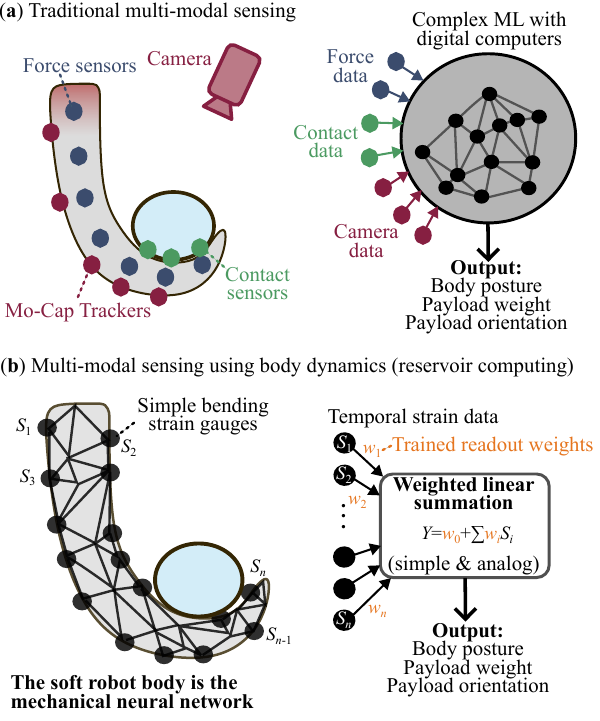}
    \caption{Multi-modal sensing strategies with traditional and proposed physical computing approaches. \textbf{To estimate the body posture, payload weight, and payload orientation}, traditional multi-modal sensing will need several kinds of sensors and cameras (a). Moreover, these sensing data will typically be processed with complex matching learning algorithms using digital computers.  In our proposed method with physical reservoir computing (b), the soft body dynamics, which are captured by the simplest bending strain gauges, can be used to extract the same information, significantly reducing the overall system complexity}
    \label{fig:bigpic}
%    \vspace{-0.2in}
\end{figure}

Thanks to their inherent compliance, soft robots have shown great potential for interacting with human operators, outpacing their rigid counterparts in the field of exploration \cite{luong2019eversion}, rehabilitation \cite{sridar2018soft}, and manipulation \cite{abondance2020dexterous}. To fully materialize such potential, the soft robots must have high-fidelity sensing capabilities to obtain complex information for effective control and decision-making. For example, we have witnessed a surge in research efforts toward posture and contact force estimation in soft robots, using camera-based motion tracking system \cite{qiao2022model}, load cells \cite{tang2022learning}, and other innovative sensors \cite{gong2021soft}. Motion capture systems provide precise position feedback of the markers attached to the soft robot's body. However, deploying motion capture systems for an outdoor task is impractical due to their limited portability. Attaching load cells only to a soft robot's end-effector could lead to inaccurate force estimation, especially when the robot body interacts with unstructured environments. Innovative soft sensors can capture extra details like arc length, contact point position, and input force magnitude. By integrating multiple sensor types --- a strategy known as multimodal sensing --- and machine learning, a complete and accurate picture of the robot's working conditions could be obtained, ultimately leading to better decision-making (Figure \ref{fig:bigpic}(a)).

However, incorporating soft sensors can complicate the overall robot design and fabrication process, leading to a substantial increase in cost and occasionally a reduction in mechanical flexibility. Moreover, an extensive sensor array may generate considerable data, necessitating high computational power and potentially causing time delays. Instead of adding specialized sensors for measuring the posture or force, one can exploit the soft robot's nonlinear body dynamics --- traditionally considered an unfavorable factor in control --- for complex information perception. Such abilities could be obtained via a non-conventional computational framework with physical reservoir computing (PRC) \cite{hauser2011towards}. 

The core principle of PRC is that a nonlinear and high-dimensional physical structure can behave like a mechanical neural network (aka the ``reservoir''), projecting input information into a high-dimensional state space with rich nonlinear dynamics. In this way, one can measure these dynamic responses to extract multiple valuable information using simple sensors (in this case, the most commonly used bending strain gauges as shown in Figure \ref{fig:arm fabrication}. In addition to freeing us from using complex and specialized sensors for information perception, another critical advantage of PRC is its minimal computation cost. Training with PRC only requires a simple linear regression with an appropriate training data selection. Then, we only need to apply a weighted linear readout to the reservoir's dynamic responses to obtain the output. 

Previous applications of PRC in soft robotic systems focused on actuation pattern generation \cite{li2012behavior}, working condition identification \cite{yu2023tapered, wang2024proprioceptive} and adaptive behaviors \cite{yoshimura2024research, li2012behavior}, aiming to address challenges in controlling soft matter. Nevertheless, little attention has been paid to obtaining better perception performance with multi-modal sensing, which could provide more information critical for effective control \cite{ kawase2021pneumatic}. Therefore, in this work, we use the PRC principle and extensive experiments to show that the dynamic responses of a soft robotic arm --- which is captured by embedded bending strain gauges --- can be exploited to extract multiple pieces of complex information simultaneously, including the body posture, payload weight, and payload orientation. In other words, the soft arm works as a multi-modal sensor with embodied machine learning capability for information perception, as shown in Figure \ref{fig:bigpic}(b). The accuracy of such embodied information perception is surprisingly high because it can distinguish very similar payload setups (i.e., the same payload with different orientations or two payloads with minimal weight differences). Once the soft robotic reservoir classifies the payload setup, it can accurately predict the trajectory of its body.

In what follows, this paper briefly describes the task of this work, the experimental setup, and the detailed setup and results for three sensing tasks. We end this paper with a conclusion and a discussion on future work.

\section{Experiments}

\subsection{Task Descriptions}

% To demonstrate the multi-modal sensing capability of our robotic arm, we investigate the potential of harnessing information processing power for predicting the soft robot’s kinematic posture and external payload, both of which are critical elements for effective control.

\begin{figure}[hbtp]
    \centering  
    \includegraphics[width=\linewidth]{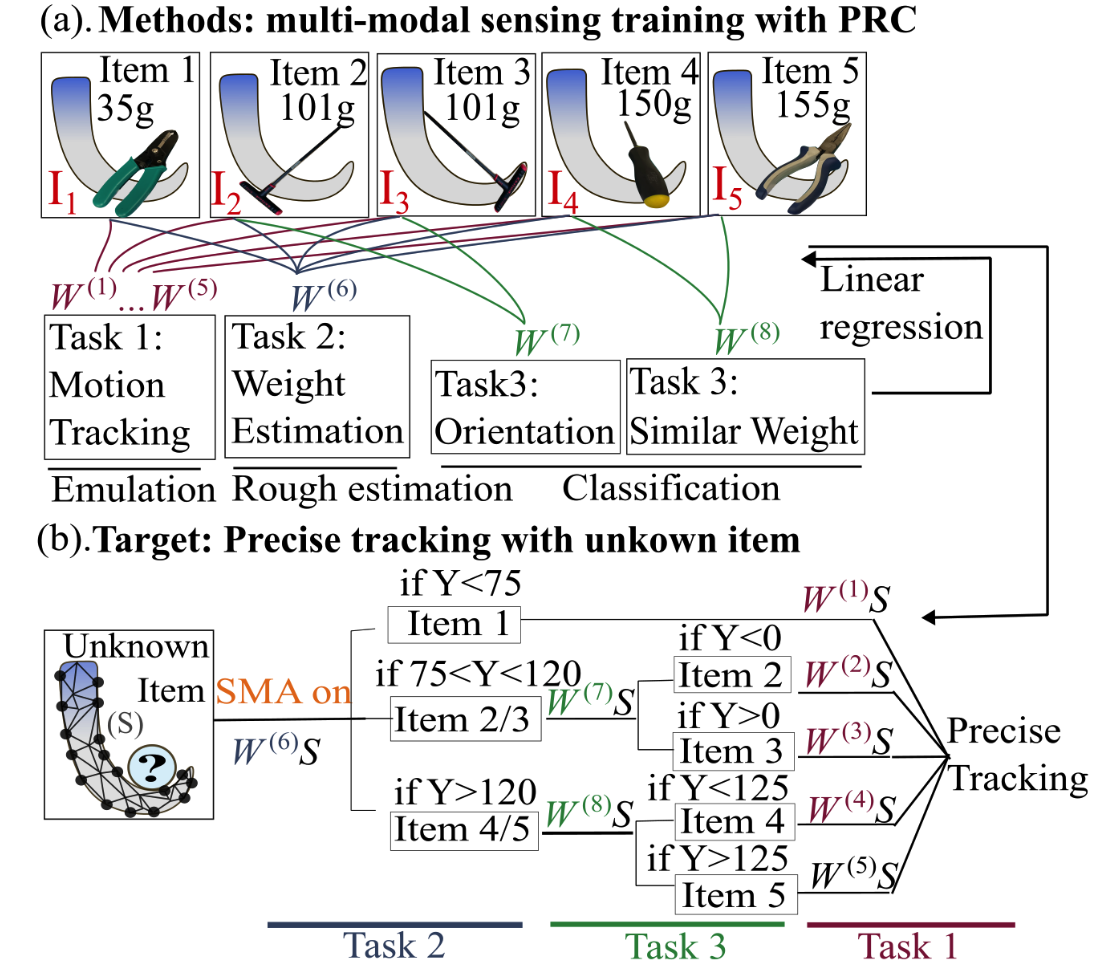}
    \caption{Schematic view of precise motion tracking with multi-modal sensing capability using physical reservoir computing. The first step is to train multiple groups of readout weights $W^{(n)}, \; n=1\ldots8$ corresponding to different information targets (a). The ultimate goal is to achieve accurate tracking regardless of the payload setup. (b) The operating procedure of the physical reservoir. }
    \label{fig:tasks}
\end{figure}

In this study, we fabricate a modular robotic arm and task it with manipulating five items typically seen in a mechanical engineer's lab: a scissor, a hammer oriented towards the left, the same hammer oriented towards the right, a screwdriver, and a plier (marked as Item 1 to 5, Figure \ref{fig:tasks}a).  Notice that items 2 and 3 are very similar in that they are the same object in different orientations.  Items 4 and 5 are also very similar because their weights only differ by 3\%.  Distinguishing between them is a challenging classification task. 

We want to train the robotic reservoir to predict body posture regardless of the payload setups. We refer to this motion-tacking task as \textit{sensing Task 1}.  
We will show it in Figure \ref{fig:training performance} that, the accuracy of Task 1's output cannot be guaranteed without knowing which item is on the soft manipulator. Therefore, we also train the robotic reservoir to identify the payload setup via two consecutive tasks -- \textit{rough weight estimation Task 2} and \textit{refined classification Task 3}. The readout weights $W^{(n)}$ associated with each task are labeled in Figure \ref{fig:tasks}(a). The methods for training these readouts are detailed afterward.

%First, the motion tracking is set as the first task for learning with arm reservoir, while arm holds different payload with griper. More specifically, 5 different payload condition are defined with common tool in mechanical engineering's lab, including scissor, hammer, screwdriver and plier, marked as item 1 to 5. 

%The most distinctive character of these payload that effect the manipulator's dynamics is the weight of the payload which is set as task 2. To demonstrate the powerful information processing capability of physical dynamics, we set two groups of payload with similar weight (Item 2 and 3 are all hammer but held by gripper in different direction, Item 4 and 5 weights are close to each other as shown in Figure\ref{fig:tasks}(a)). Task 3 is added to train the arm to identify these minor differences with high-dimensional deformations of its movement. Readout weights for three tasks (W1 to W7) could be trained in parallel by simple linear regression after collecting 5 groups of dynamic state vectors under each payload condition.

In what follows, we describe how this robot operates (Figure\ref{fig:tasks}(b) and supplement video). First, we use the embedded SMA coil to actuate the robotic arm. The response of such SMA actuation is dynamic: the robotic arm shows a short burst of transient vibration before settling into the slower bending motion.  Denote state vector matrix $S$ as the reading from the embedded bending gauge sensors. We can apply readouts $W^{(6)}$ to $S$ to estimate the payload weight ($Y$), dividing the payload setup into three scenarios (aka., Sensing Task 2): If the payload weight is roughly below 75 grams, Item 1 is on the robot. If the payload is between 75 and 120 grams, Item 2 or 3 is on the robot. If the weight is above 120 grams, Item 4 or 5 is on the robot. 

If payload weight estimation indicates the second or third scenarios, we will apply refined classification Task 3. That is, we will apply readout $W^{(7)}$ to $S$ and distinguish between the hammer's orientation, or apply $W^{(8)}$ to differentiate between 150 grams screwdriver and 155 grams plier. Once the payload items are identified, we can apply readout weights $W^{(1)}$ to $W^{(5)}$ correspondingly to predict the robotic motion. 

The primary research questions answered in this work are threefold: First, does the time-series data from the embedded sensor network have enough dynamics to estimate the robotic movement? Second, could the arm accurately differentiate similar payloads even with minor differences? Third, are embedded bending sensors the preferred option for capturing body dynamics and reservoir computing, compared to other approaches like the traditional camera?

\subsection{Experimental Setup}

\subsubsection{Fabrication of the Soft Modular Arm}

\noindent\par
\begin{figure}[htb]
    \centering  
    \includegraphics[width=\linewidth]{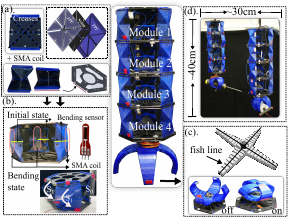}
    \caption{Fabrication of the soft modular robotic arm with four modules. The module manufacturing process consists of two steps: (a) Panel fabrication and (b) assembly with the bending sensor and shape memory alloy. (c) A 3D-printed gripper is attached to the four connected modules. (d) The arm's size can be scaled up or down.}
    \label{fig:arm fabrication}
\end{figure}

The robotic arm in our study (Figure \ref{fig:arm fabrication}) consists of four modules and one gripper, all fabricated with 3D-printed components for an easy manufacturing and assembly process. Each elementary module comprises two rigid PLA plates connected by three foldable soft panels (Figure \ref{fig:arm fabrication}(a)). The main component of each lateral panel is a sandwich structure (TPU, nylon,TPU) inspired by the Yoshimura design \cite{deshpande2024golden}, as illustrated in Figure \ref{fig:arm fabrication}(a). These panels can flexibly fold along predefined creases (dashed lines in Figure \ref{fig:arm fabrication}(a)), allowing controlled deformation.

Each panel is equipped with shape memory alloy (SMA) coils (Kellogg's Research Labs) and a one-inch bending sensor (Flexpoint Sensor Systems, Inc.), both placed along the horizontal bending crease (Figure \ref{fig:arm fabrication}(b)). The folding panels can be assembled onto the base plate like LEGO component, eliminating the need for additional assembly hardware. At the end of the arm, a 3D-printed gripper with central alignment holes in four directions is attached to the last module. A fishing line is threaded through the center of each finger from the outside to the center. When the servo motor (MG90D, Tower Pro Ltd.) rotates 60 degree via an Arduino micro-controller, the gripper closes to grasp the payload, as illustrated in Figure \ref{fig:arm fabrication}(c). 

For all experiments, the panels of the four modules are reset to an inward-folded configuration before testing (Figure \ref{fig:arm fabrication}). This ensures a consistent initial condition, which is crucial for maintaining the performance of the reservoir computing framework. During testing, two panels transition to the bending state when the SMA is powered. The flexible panels fold along the central crease without internal facet bending (Figure \ref{fig:arm_reservoir}), and the resulting sensor dynamics are utilized for multi-sensing tasks. Moreover, this flexible, modular, and easy-to-assemble design facilitates scalability, as demonstrated in Figure \ref{fig:arm fabrication}(d).

\subsubsection{Actuation and Sensor Configuration}
\vspace{6pt}
\noindent\par

\begin{figure}[hbtp]
    \centering  
    \includegraphics[width= 0.95\linewidth]{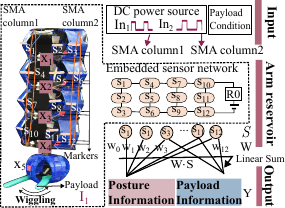}
    \caption{Schematic view of precise motion tracking with the arm's multi-modal sensing capability enabled by reservoir computing. The arm bends after two SMA columns are heated. The wiggling behavior of arm is captured by sensor network, which serves as the computing kernel for kinematic and payload information perception.}
    \label{fig:arm_reservoir}
\end{figure}

All SMAs in the same column are series-connected, and two such columns are actuated by separate PWM signals—denoted as $In_{1}$ and $In_{2}$ —with a phase difference of 0.2 seconds. Each actuation cycle consists of 0.2 seconds of heating followed by 0.4 seconds of relaxation to prevent overheating. This actuation mechanism is illustrated in Figure~\ref{fig:arm_reservoir}. Each SMA column is powered by a 30V/6A DC power supply, and the actuation signals are controlled via an Arduino Uno.

The sensor network is wired such that sensors within each module are series-connected, while sensors across different modules are parallel-connected, as shown in Figure~\ref{fig:arm_reservoir}. The entire sensor system is connected to a 2k$\Omega$ resistor and a 6V DC power supply (Figure~\ref{fig:arm_reservoir}). Voltage readings from each sensor are collected and transmitted to a host PC every 1 millisecond via a 16-channel data acquisition module (NI 9205, National Instruments Corp.). The sensor voltage is positively correlated with the bending angle, aligning with the creases in each module where the sensors are attached. This 12-sensor network effectively captures the deformation of the 12-panel structure, providing a comprehensive representation of the overall dynamic behavior. For example, when SMA column 1 is actuated, the voltage across sensors S1, S4, S7, and S10 should increase, while all others should decrease. The nonlinear nature of the sensor network emerges from two manifolds: 1) The intrinsic material and geometric nonlinearity of the robotic arm movement.2) The nonlinear bending-resistance response of the sensors. 

The posture of the arm is recorded by a camera (Canon EOS R3, 60 fps) through nine red markers (two for each module and one for the gripper) during all experiments. The horizontal displacement of the center of each module is calculated as the average position of the corresponding markers, labeled as \( x_1, \ldots, x_4 \). The motion of the gripper is labeled as \( x_5 \). These kinematic data serve as the target output when training the arm for the kinematic perception task. After training, the arm should be able to perform three sensing tasks without relying on visual data. However, visual data are still utilized to evaluate the test performance.

Additionally, visual data are used as computational state vectors for the payload information perception task, providing a comparison to the embodied sensor data. Since visual data also capture high-dimensional system dynamics, they can serve as a computational resource \cite{jun2023PRC}. This cross-comparison offers insights into perception training using different information sources and helps verify the effectiveness of the sensor network in capturing the arm's dynamic behavior.

\subsubsection{Framework of Multi-modal Sensing with Modular Arm Reservoir}
\vspace{6pt} 
\noindent\par

Figure~\ref{fig:arm_reservoir} illustrates the information processing flow in the modular arm. The actuation signal and payload condition both serve as the input layer for the arm reservoir, as they collectively define the dynamic response characteristics. After initiating SMA actuation, each module is manually set to a flexible state, causing the arm to exhibit transient bending movements followed by small oscillations around a stable bending angle. Each experiment runs for 30 seconds.

Data from the twelve bending sensors, denoted as $S = [s_1, s_2, \dots, s_{12}]^\intercal$, and nine visual data points are recorded simultaneously. To synchronize the sampling frequencies between the data acquisition system (DAQ) and the camera for posture training, 20 frames per second are selected from the camera (out of 60 fps), and 20 frames per second are selected from the DAQs (out of 1000 fps). The resulting compacted dataset $S$ constitutes the dynamic state vectors. The experiment is repeated 10 times for each payload condition $(I_1, \dots, I_5)$. These 5 $\times$ 10  groups of data (600 frames per dataset) serve as the training and testing datasets. During information processing, the actuated robotic body functions as a fixed reservoir kernel.

Training the arm reservoir for all sensing tasks follows the same process: determining different sets of static linear readout weights (\( W^{(1)}, \dots, W^{(8)} \)) using the collected state vectors and target functions \( Y = \hat{y(t)} \). Once the readout weights are obtained via linear regression during training, the reservoir output during testing is computed as $y(t) = w_0 + \sum\limits_{i=1}^{12} w_i s_i.$

The accuracy of these predictions is evaluated using the normalized root mean squared error (NRMSE) to ensure a fair comparison across different tasks. The NRMSE is defined as ${NRMSE} = \frac{\text{RMSE}}{\text{Mean of actual values}} = \frac{\sqrt{\frac{1}{N} \sum_{i=1}^{N} (y_i - \hat{y}_i)^2}}{\frac{1}{N} \sum_{i=1}^{N} y_i}, $where \( N \) is the number of time steps in the testing period, \( y_i \) is the reservoir output at the \( i \)th step, and \( \hat{y}_i \) is the true value at the \( i \)th step.

A detailed explanation of the training targets and experimental setup is provided for each task.

\subsection{Sensing task 1: posture information}

The first task explores whether the robotic arm can learn its motion patterns using its embedded sensing system. Five different payload conditions represent distinct input scenarios.

\begin{figure}[hbtp]
    \centering  
    \includegraphics[width=\linewidth]{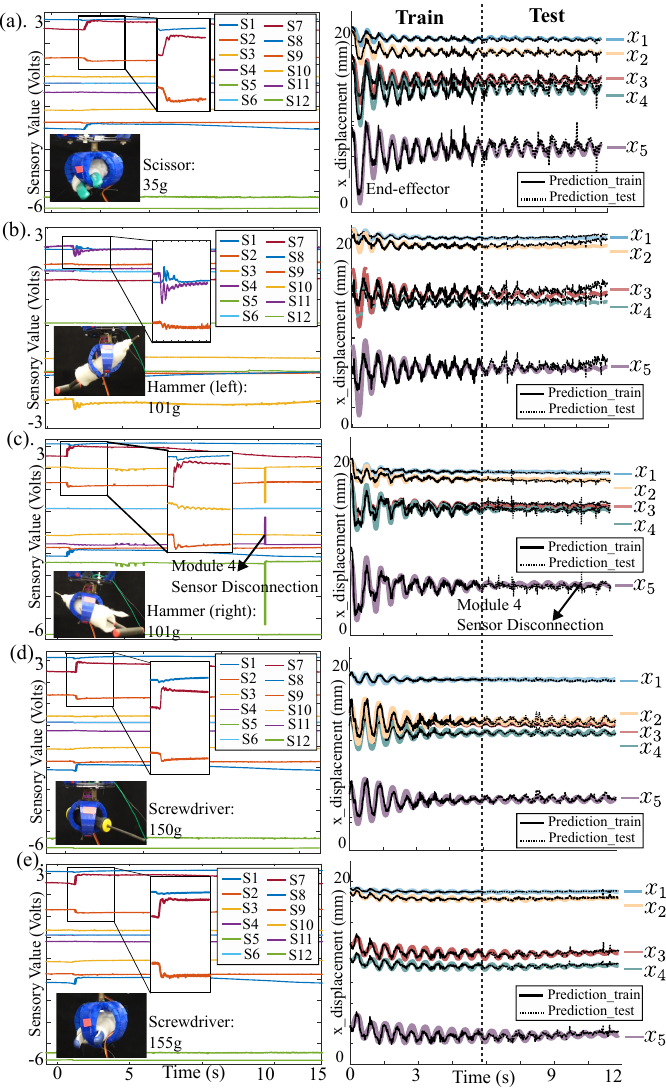}
    \caption{An example of motion prediction (horizontal displacement) using the robotic arm's sensor network. (a) represents the collected sensor data, while (b) shows the corresponding training and testing results. The five colorful solid lines indicate the actual movements of module 1 through module 4 and the gripper. The first half of the black solid line represents the reservoir output of the arm, corresponding to five tracked points, while the latter half of the dashed line represents the predicted motion during the testing phase.}
    \label{fig:motion tracking}
\end{figure}

 Under each condition, the reservoir output is expected to predict the arm's movements, represented by five tracked points (i.e., \( X(t+1) \)), using the current system state $S(t)$. Consequently, the target function for training is defined as $x(t+1)=w_0+\sum\limits_{i=1}^{12} w_i s_i(t)$, where five sets of readout weights for gripper's tracking (\( W^{(1)}, \dots, W^{(5)} \)) are trained under the five payload conditions using linear regression (LR). In each dataset, the first 6 seconds (120 frames) are used for training, while the following 120 frames are reserved for testing.

Figures \ref{fig:motion tracking}(a) to (e) illustrate example outputs of the training and testing processes for motion tracking under the five payload conditions, along with the corresponding twelve sensor readings. The five colorful lines in Figures \ref{fig:motion tracking}(f) to (j) represent the actual trajectories of five tracked points, while the solid and dashed black lines correspond to the training and prediction results, respectively. The results demonstrate that the embedded sensory system can effectively predict the robotic arm’s motion across all cases. Although the sensor values do not exhibit an apparent fluctuation pattern similar to the recorded displacement trajectory, the sensor dynamics still encode sufficient motion information. The vibration pattern is visible in the enlarged sensor data, as shown in Figure \ref{fig:motion tracking}. 

Some noise is observed in the testing phase of reservoir output, which arises from high-frequency fluctuations in sensor data and occasional sensor disconnections. Nevertheless, the arm reservoir exhibits robustness to such noise. For instance, the disconnection of a sensor in module four caused the voltage across $S_{10}$ and $S_{11}$ to drop to zero while $S_{12}$ remained equal to the power supply voltage (6V). However, as shown in Figure \ref{fig:motion tracking}(h), this did not cause a significant deviation in motion prediction. This robustness is likely due to the high dimensionality of the sensor data, which allows the system to compensate for missing or noisy signals.

\begin{figure}[hbtp]
    \centering  
    \includegraphics[width = \linewidth]{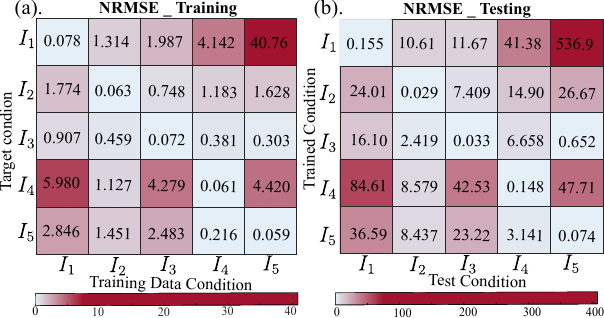}
    \caption{(a) Averaged trianing NRMSE when training state vectors and target are only matched in diagonal. (b) Averaged NRMSE test results for the trained arm is tested under different payload conditions.}
    \label{fig:training performance}
\end{figure}

Figure \ref{fig:training performance}(a) presents the training performance of the gripper's trajectory tracking (i.e., $x_5$) using the averaged NRMSE. The rows and columns correspond to the payload conditions (\( I_1, \dots, I_5 \)) under which the sensor data and target displacements were collected. Only in the diagonal cases, where the sensor and target data were collected under the same payload condition, is the training error significantly lower. This indicates substantial variations in arm dynamics across different payloads, and the sensor network effectively captures these variations for training.

Figure \ref{fig:training performance}(b) shows the averaged NRMSE results for gripper movement estimation using five different trained models ($I_1$ to $I_5$). The RMSE remains significantly lower only when the training and testing data correspond to the same payload condition. Even though conditions 2/4 and 3/5 have similar payload weights, their trained readout weights are not interchangeable. It highlights the importance of accurately estimating the payload before achieving precise motion tracking for control tasks. Such estimation can be accomplished in two steps, as discussed in the subsequent sections (tasks 2 and 3).

\subsection{Sensing task 2: Rough Payload information - weight estimation}

\begin{figure}[hbtp]
    \centering  
    \includegraphics[width=\linewidth]{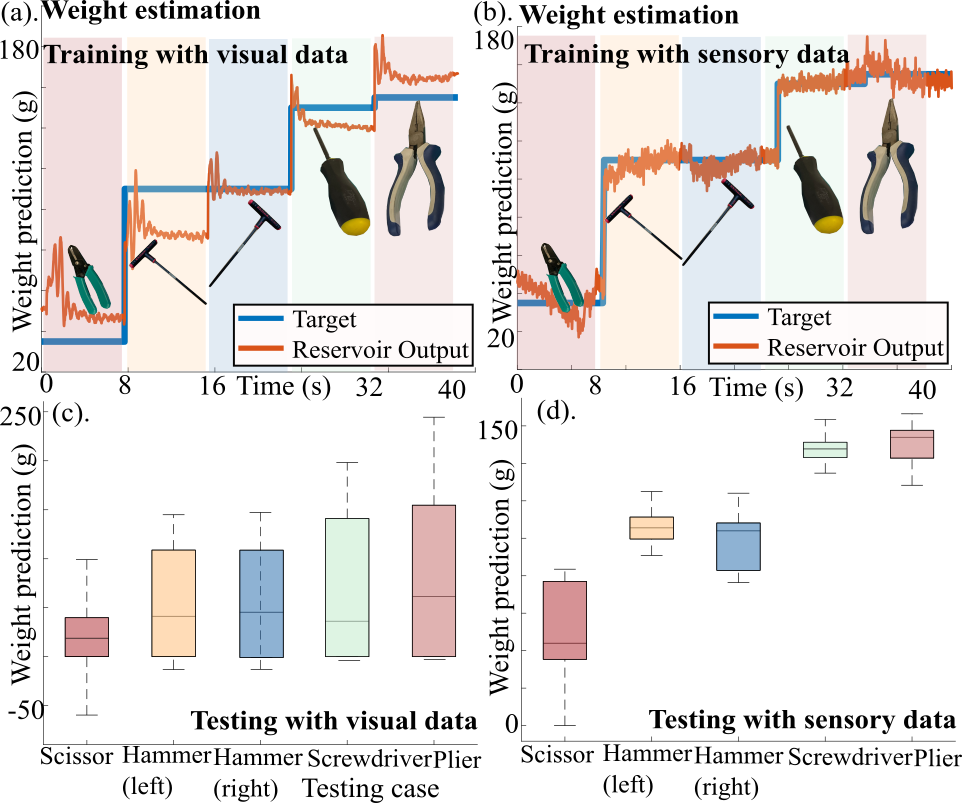}
    \caption{Results of weight estimation sensing task. (a) and (b) are training results with visual and sensor data respectively. (c) and (d) are corresponding testing results with the trained arm.}
    \label{fig:weight estimation}
\end{figure}

Task 2 aims to perform a rough classification of all payload conditions by setting the payload weight as the corresponding target function during training. Additionally, we compare the learning capability of the robotic arm using both visual data and embedded sensor data. During training, we randomly select one vibration test from ten repeated trials under each payload condition and extract the first 8 seconds of data (160 frames) from five datasets as training data. Then, five groups of sensor (12 bending sensor)/displacement (9 tracked points) data obtained from DAQs/cameras are compiled into state vectors, defined as $S(t)=[S_{I1};S_{I2};S_{I3};S_{I4}; S_{I5}] \quad \text{or} \quad S_2(t)=[X_{I1};X_{I2};X_{I3};X_{I4}; X_{I5}]$, where $S_{I1}$ and $X_{I1}$ are sensor data and visual data collected with Item 1.  And the corresponding target function is the payload weight $Y(t)=[m_1;m_2;m_3;m_4;m_5]$. This piecewise training function is shown as the blue line in Figure \ref{fig:weight estimation}(a) and (b). The training results obtained using visual data and the sensor network are plotted as orange lines in Figures \ref{fig:weight estimation}(a) and (b), respectively. Although the visual data contain less noise compared to the bending sensor data, the training output using the bending sensor is closer to the target function. This suggests that the bending sensor network captures more distinct characteristics of the dynamic behaviors, offering better separability than visual data-based training.

For testing, we randomly select 8 seconds of data from the total dataset (60 s $\times$ 10 trials) for each payload condition. The estimated payload weight is computed as the average value of the weighted linear summation over 480 frames. Testing for each payload condition is repeated 50 times, and the estimation results are summarized in Figure \ref{fig:weight estimation}, where (a) represents weight estimation using visual data and (b) represents weight estimation using sensor data. 

Each box in the figure represents the distribution of estimated weights, where the bottom and top edges correspond to the 25th and 75th percentiles, respectively. The results clearly show that the visual data fails to separate the five payload types due to limited dimensionality. The five groups of estimated weights significantly overlap, showing large variations. In contrast, training with sensor data effectively distinguishes all test results into three distinct groups with smaller deviations. This indicates that the sensor network allows the arm to be well-trained for payload weight estimation. Another possible reason for the smaller deviation in the sensor-based estimation is the bending-voltage relationship. Small posture variations during testing do not result in significant deviations in sensor readings, making the estimation more stable.

\subsection{Sensing task 3: Precise Payload information - further classification}

\begin{figure}[hbtp]
    \centering  
    \includegraphics[width=0.8\linewidth]{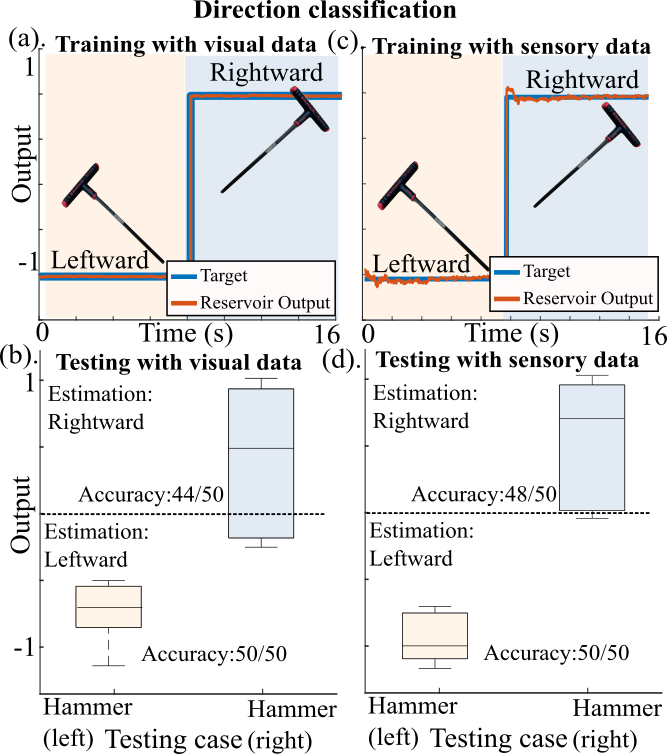}
    \caption{Classification results when hammer in different direction. (a) and (b) are training results with visual and sensor data respectively. (c) and (d) are corresponding testing results with the trained arm.}
    \label{fig:direction}
\end{figure}

Task 3 aims to further distinguish payloads with similar weights, such as a hammer held in different orientations and a screwdriver/plier with nearly identical weights. Similar to the previous tasks, both visual data and embedded sensor data are utilized. For direction classification training, the state vector is compiled using data collected from items 2 and 3, defined $S(t)=[S_{I2};S_{I3}] \quad \text{or} \quad S(t)=[X_{I2};X_{I3}]$, with the corresponding target function $Y(t)=[-1;1]$ as shown in Figure \ref{fig:direction}. The testing procedure follows the same methodology as in Task 2, with 50 repeated trials, where the average value of the weighted linear summation is taken as the arm's prediction. If the output value is less than zero, the prediction is classified as left-facing; otherwise, it is classified as right-facing. Both visual and sensor-based approaches exhibit high prediction accuracy, achieving 88\% and  96\%, respectively.

\begin{figure}[hbtp]
    \centering  
    \includegraphics[width=0.8\linewidth]{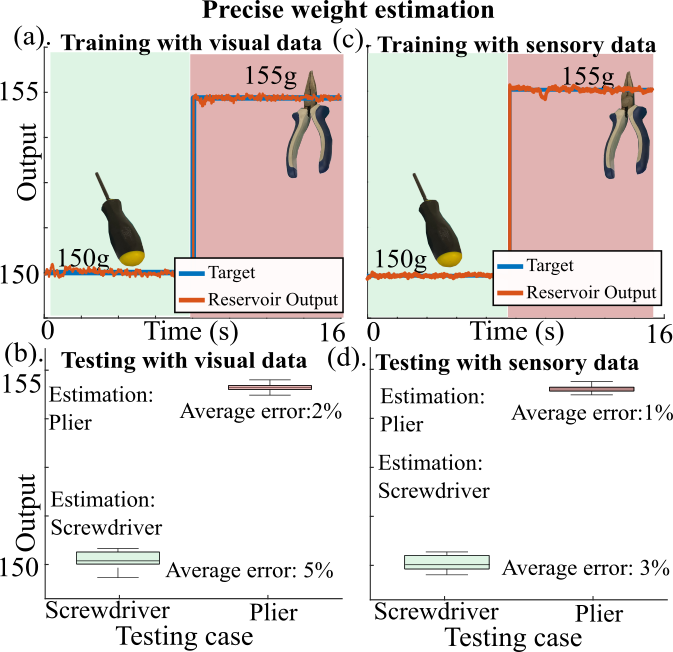}
    \caption{Results of classification of payloads with close weight. (a) and (b) are training results with visual and sensor data respectively. (c) and (d) are corresponding testing results with the trained arm.}
    \label{fig:closeweight}
\end{figure}

For payload classification of objects with similar weight, the state vector is compiled using data collected from items 4 and 5 $S(t)=[S_{I4};S_{I5}] \quad \text{or} \quad S(t)=[X_{I4};X_{I5}]$ with the corresponding target function Y(t)=[150; 155] as shown in Figure \ref{fig:closeweight}. Similar to the previous classification task, both visual and sensor-based approaches achieve high prediction accuracy, with an error margin of less than 5\%.

These results indicate that while displacement dynamics alone may not be sufficient for training a perception model across all five payload conditions, they still encode valuable information that enhances the robotic arm's physical perception capabilities within specific constraints.

\subsection{sensing capability with reduced sensor}

\begin{figure}[hbtp]
    \centering  
    \includegraphics[width=\linewidth]{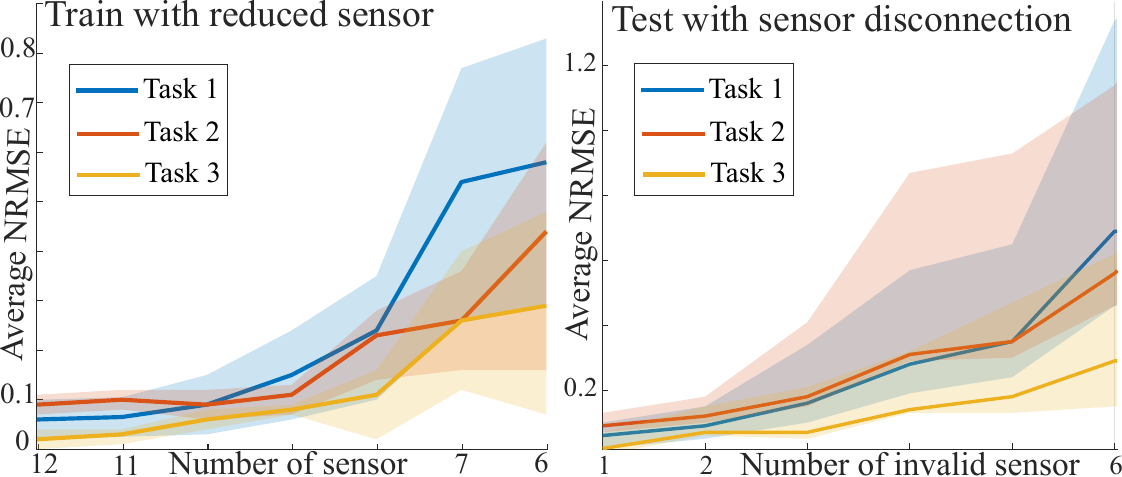}
    \caption{(a) Averaged RMSE of each task over all payload conditions when the number of available sensors is reduced during both training and testing. (b) Averaged RMSE for each task when a subset of sensors malfunctions during testing, while the arm has been trained with all sensors. The shaded area represents the standard deviation in each scenario.}
    \label{fig:sensoranalysis}
\end{figure}

Figure \ref{fig:sensoranalysis}(a) presents the testing results for the three tasks when the number of available sensors is reduced. Performance is evaluated using the averaged NRMSE across all trials and payload conditions. The error remains well-controlled within 10\% when more than ten sensors are available during both training and testing. This result highlights the importance of the high-dimensional nature of sensor data in enhancing perception capability.

Figure \ref{fig:sensoranalysis}(b) illustrates the averaged NRMSE when a subset of sensors malfunctions during the testing phase. This scenario is simulated by enforcing the values of selected sensors to zero only during testing. Such an analysis is critical for real-world applications, as it assesses the robustness of the perception system when encountering sensor disconnection issues. The results indicate that the error remains relatively low when only one or two sensors fail during testing. However, as more sensors become non-functional, the deviation in error increases significantly. This suggests that the arm can still maintain adequate sensing capability if the removed sensors are not positioned in critical locations. Future work will focus on optimizing sensor distribution to enhance system resilience against sensor failures.

\section{CONCLUSIONS}

 In this work, we demonstrated the multi-information perception capability of a modular soft robotic arm using the physical reservoir computing framework. The embedded bending sensor network  could efficiently capture the characteristics of robotic dynamics to predict both its posture (node tracking) and the accurate payload information (weight, direction) simultaneously. More importantly, accurate real-time tracking with varied working conditions is possible with the proposed methods, by adaptively selecting the appropriate stored trained readout weight after identifying the payload information. This study paves the way for future developments in adaptive, real-time control of soft robots using embedded intelligence, reducing reliance on external sensors and complex computational resources. For future work, we aim to further integrate this multi-modal sensing capability into adaptive closed-loop control of the robotic arm . Besides, we plan to optimize the system by refining the structural design and sensor distribution to enhance the inherent computing capacity of the robotic reservoir. These improvements will empower soft robots with more effective and intelligent perception, feedback control, and adaptability for advanced robotic applications.

\addtolength{\textheight}{-12cm}   % This command serves to balance the column lengths

%%%%%%%%%%%%%%%%%%%%%%%%%%%%%%%%%%%%%%%%%%%%%%%%%%%%%%%%%%%%%%%%%%%%%%%%%%%%%%%%

%%%%%%%%%%%%%%%%%%%%%%%%%%%%%%%%%%%%%%%%%%%%%%%%%%%%%%%%%%%%%%%%%%%%%%%%%%%%%%%%

%%%%%%%%%%%%%%%%%%%%%%%%%%%%%%%%%%%%%%%%%%%%%%%%%%%%%%%%%%%%%%%%%%%%%%%%%%%%%%%%

\section*{ACKNOWLEDGMENT}

J. Wang and S. Li acknowledge the support from National Science Foundation (CMMI-2239673,CMMI-2328522) and Virginia Tech (via startup funds).

%\section*{References}
\bibliographystyle{IEEEtran}
\bibliography{ref}

\end{document}